\documentclass[conference]{IEEEtran}
\usepackage[compatibility=false]{caption}

\IEEEoverridecommandlockouts
\usepackage{hyperref}
\usepackage{amsmath,amssymb,amsfonts}
\usepackage{graphicx}
\usepackage{textcomp}
\usepackage{xcolor}
\usepackage{cite}
\usepackage{circuitikz}
\usepackage{tikz}
\usepackage{subcaption}
\usetikzlibrary{positioning}

\UseRawInputEncoding

\begin{document}

\title{High-Resource Translation: \\ 
Turning Abundance into Accessibility}

\author{
    \IEEEauthorblockN{
        Yanampally Abhiram Reddy\IEEEauthorrefmark{1}
    }
    \IEEEauthorblockA{\IEEEauthorrefmark{1}ABV-IIITM Gwalior, MP, India}
    
    }

\maketitle

\begin{abstract}
This paper presents a novel approach to constructing an English-to-Telugu translation model by leveraging transfer learning techniques and addressing the challenges associated with low-resource languages. Utilizing the Bharat Parallel Corpus Collection (BPCC) as the primary dataset, the model incorporates iterative backtranslation to generate synthetic parallel data, effectively augmenting the training dataset and enhancing the model's translation capabilities. 

The focus of this research extends beyond mere translation accuracy; it encompasses a comprehensive strategy for improving model performance through data augmentation, optimization of training parameters, and the effective utilization of pre-trained models. By adopting these methodologies, we aim to create a more robust translation system that can handle a diverse range of sentence structures and linguistic nuances inherent to both English and Telugu. 

This research highlights the significance of innovative data handling techniques and the potential of transfer learning in overcoming the limitations posed by sparse datasets in low-resource languages.This research not only contributes to the field of machine translation but also aims to facilitate better communication and understanding between English and Telugu speakers in real-world contexts. Future work will concentrate on further enhancing
the model’s robustness and expanding its applicability to more
complex sentence structures, ultimately ensuring its practical
usability across various domains and applications. 
\end{abstract}


\section{Introduction}

Machine translation (MT) is a significant subfield of natural language processing (NLP) that focuses on automatically translating text from one language to another. As the world becomes more interconnected, the need for reliable translation tools across various languages is increasing. While some languages, like English, benefit from large amounts of parallel text and established models, many regional and low-resource languages, such as Telugu, face challenges due to limited training data \cite{gala2023indictrans}. Developing effective MT systems for these languages is crucial for breaking down communication barriers, preserving cultural diversity, and enabling access to information.

This project focuses on enhancing the translation of English to Telugu, a Dravidian language spoken by millions, by utilizing modern transfer learning techniques and backtranslation to overcome the limitations posed by the scarcity of parallel data \cite{gala2023indictrans}. Traditional MT approaches like rule-based or statistical methods require substantial linguistic knowledge or large bilingual corpora, making them less effective for low-resource languages. The introduction of neural machine translation (NMT) has revolutionized MT by leveraging deep learning to model complex language patterns \cite{bahdanau2014}. However, NMT models require large amounts of parallel data, which remains a challenge for languages like Telugu.

To address the challenge of limited data for machine translation, we employ transfer learning, a technique that allows us to adapt a pre-trained model from a large multilingual corpus to a specific language pair, in this case, English-Telugu \cite{gala2023indictrans}. In this project, we begin by fine-tuning the \textbf{Helsinki-NLP/opus-mt-tc-big-en-it} model, which was originally trained on high-resource languages like English and Italian. One key reason for selecting an English-Italian model is the notable phonetic and rhythmic similarities between Italian and Telugu, often described in terms of their “musicality.” This linguistic quality is also why Telugu has historically been referred to as the \textbf{“Italian of the East.”} Both languages have a melodious nature, characterized by a rich vowel system, consistent phonetic rules, and a balanced syllabic structure \cite{gala2023indictrans}. These similarities enhance the effectiveness of transfer learning when adapting an Italian-based model to Telugu.

By fine-tuning this model on our English-Telugu parallel dataset \cite{gala2023indictrans}, we can leverage the general language patterns and knowledge embedded in the pre-trained model and transfer it to Telugu, even with a relatively small amount of training data. This method allows us to achieve higher performance in translation tasks by building upon the pre-existing linguistic relationships, rather than starting from scratch with Telugu, which is considered a low-resource language in the NLP landscape.

The tokenization process is crucial to achieving good performance in NMT, especially when working with morphologically rich languages like Telugu. We use \textbf{SentencePiece}\cite{kudo2018sentencepiece}, a data-driven approach to subword tokenization that allows us to handle out-of-vocabulary words effectively \cite{sennrich2016}. SentencePiece generates a subword vocabulary from both English and Telugu sentences, ensuring that even rare words or complex morphology can be tokenized and understood by the model.

A key component of this project is the use of backtranslation, a technique widely used to improve the performance of NMT models in low-resource settings \cite{edunov2018}. Backtranslation involves generating synthetic parallel data by translating monolingual target-language text (Telugu) back into the source language (English). This approach augments the training dataset and provides the model with additional examples of how the two languages relate, thus improving its ability to translate in both directions. Iterative backtranslation , which involves multiple cycles of generating synthetic data and retraining the model, allows the system to continuously improve its understanding of the target language.

In summary, this project demonstrates how transfer learning and backtranslation can be combined to build a robust machine translation system for English and Telugu, despite the scarcity of parallel data. By leveraging state-of-the-art NMT models \cite{vaswani2017}, efficient tokenization \cite{sennrich2016}, and iterative training techniques \cite{edunov2018}, we aim to improve translation quality and contribute to the development of tools for low-resource languages like Telugu. The resulting model can support a variety of applications, including multilingual communication, cultural preservation, and information accessibility for Telugu speakers.

\section{Related Works}

    Backtranslation has been widely used in machine translation tasks, especially for low-resource languages. Research has shown that generating synthetic data using backtranslation can improve the quality of translation models by compensating for the lack of parallel data. Previous work includes applications in high-resource languages like English-French and English-German, but its application to low-resource languages like Telugu is less explored \cite{bawden-etal-2020-university}.
    
    In their work, Bawden et al. (2020)\cite{bawden-etal-2020-university} describe the University of Edinburgh's submissions to the WMT20 news translation shared task, focusing on two language pairs: English-Tamil, which is considered very low-resource, and English-Inuktitut, a mid-resource pair. The authors utilized the neural machine translation transformer architecture for their submissions and implemented various techniques aimed at enhancing translation quality in the face of limited parallel training data. For the English-Tamil pair, they explored methods such as pretraining, leveraging language model objectives, and utilizing an unrelated high-resource language pair (German-English) to facilitate translation through iterative backtranslation. Furthermore, for English-Inuktitut, the researchers investigated the use of multilingual systems, which, although not part of the primary submission, showed promising results on the test set. This work highlights the potential of backtranslation and other innovative strategies in improving translation models for low-resource languages.

\section{Methodology}

This section outlines the core methodology followed to enhance the performance of the English-Telugu machine translation model, particularly focusing on techniques like transfer learning, backtranslation, and iterative fine-tuning. Each step in the methodology is directly linked to key functions and operations that execute the translation pipeline, ensuring replicability and clarity of the process.

\subsection{Data Preparation}

Machine translation models require a bilingual corpus for training. However, due to the limited availability of English-Telugu parallel data, data preparation becomes a critical step in improving model performance. The initial dataset used for training consists of English-Telugu sentence pairs sourced from the \textbf{Bharat Parallel Corpus Collection (BPCC)} \cite{gala2023indictrans}, which provides a valuable resource for building robust translation models in low-resource language settings.

\subsubsection{Training SentencePiece Model}
To handle the diversity of linguistic forms in both languages and to efficiently tokenize the text, a subword tokenization approach is adopted using SentencePiece. SentencePiece \cite{kudo2018sentencepiece} is an unsupervised text tokenizer and detokenizer designed to facilitate neural network-based language processing tasks, particularly for languages with rich morphology and varying script systems. 

\subsubsection{Overview of SentencePiece}
SentencePiece \cite{kudo2018sentencepiece} works by segmenting text into subword units rather than relying on word-level tokenization. This approach helps manage the vast vocabulary size and out-of-vocabulary (OOV) words that can occur in low-resource languages like Telugu. The primary advantages of using SentencePiece include:

\textbf{1. Handling Rare Words:} By breaking down words into smaller subword units, SentencePiece can represent rare or unseen words as combinations of more common subwords, improving translation robustness.

\textbf{2. Language Independence:} SentencePiece operates independently of the language, making it suitable for multilingual tasks. It generates a vocabulary based solely on the statistical distribution of subwords within the provided training data.

\textbf{3. Subword Regularization:} The model can utilize different variations of subword tokenization during training to enhance robustness. This means it can capture various morphological forms and syntactic structures effectively.

\subsubsection{\textbf{Training Process}}
The training of the SentencePiece model combines English and Telugu sentences to generate a shared vocabulary. The following steps outline the training process:

\textbf{1. Data Preparation:} The first step involves collecting and preprocessing the English and Telugu sentence pairs. This includes normalizing the text, removing unwanted characters, and ensuring consistency in encoding.

\textbf{2. Tokenization Algorithm:} SentencePiece employs a Byte Pair Encoding (BPE) or Unigram Language Model (ULM) approach to determine the optimal segmentation of text into subwords. In this implementation, we specifically utilize the BPE method, as it is particularly effective for the characteristics of our bilingual dataset and the requirements of the translation task.

\subsubsection{\textbf{Byte Pair Encoding (BPE)}}
Byte Pair Encoding (BPE) iteratively replaces the most frequent pair of bytes in the dataset with a new token. This process continues until the desired vocabulary size is achieved. The mathematical representation of BPE can be summarized as follows:
\[
\text{merge}(x, y) = \arg\max_{(x,y) \in S} f(xy) 
\]
where \( S \) is the set of all byte pairs, and \( f(xy) \) represents the frequency count of the merged pair \cite{sennrich2015}.

\subsubsection{\textbf{Unigram Language Model (ULM)}}
In contrast, the Unigram Language Model (ULM) treats subword units as a probabilistic model, where the likelihood of a sentence can be expressed as:
\[
P(S) = \prod_{i=1}^{n} P(w_i | S_{<i})
\]
In this equation, \( S \) denotes the sentence, \( w_i \) represents the tokens, and \( S_{<i} \) are the preceding tokens.

\textbf{3. Vocabulary Size and Model Type:} The model is trained using essential parameters such as \texttt{vocab\_size} and \texttt{model\_type}. The \texttt{vocab\_size} parameter determines the maximum number of subword units that the model can learn, which is typically set based on the characteristics of the dataset and the linguistic properties of the languages involved. In our implementation, we have set the \texttt{vocab\_size} to \textbf{16,000}. This choice reflects a careful consideration of the dataset's diversity and the need to balance sufficient linguistic variation with efficient processing capabilities.

The \texttt{model\_type} parameter specifies the tokenization method to be employed, either Byte Pair Encoding (BPE) or Unigram Language Model (ULM). In this case, we adopted BPE for our SentencePiece model. This choice is particularly advantageous, as BPE efficiently handles the linguistic diversity inherent in both English and Telugu. By using BPE, we enhance the model's ability to generate subword units that accurately represent the syntactic and semantic structures of both languages, thereby improving the overall performance of our translation system.

\textbf{4. Training the SentencePiece Model:} The model is trained on the combined English and Telugu datasets. During this phase, it learns to tokenize sentences into subwords, capturing the distribution and structure of both languages effectively.

\textbf{5. Applying the Tokenizer:} Once the SentencePiece model is trained, it can be applied to the English and Telugu text, converting sentences into sequences of subword tokens. This representation facilitates efficient processing and improves the performance of downstream tasks like machine translation.

In summary, the SentencePiece model provides an effective way to tokenize bilingual datasets by leveraging subword units, allowing for better handling of linguistic diversity and enhancing the overall performance of the translation system.

\subsubsection{\textbf{Tokenization with SentencePiece}}
The trained SentencePiece model is used to tokenize both English and Telugu sentences into subword units. Tokenization functions are implemented to encode sentences into subword sequences, preparing the dataset for training a transformer model. Tokenized datasets include both input token IDs and attention masks for further model processing.

\subsection{Transfer Learning}

Due to the scarcity of large-scale parallel corpora for English-Telugu translation, we employ transfer learning. This technique allows us to utilize a pre-trained model that has been trained on a high-resource language pair and fine-tune it for our low-resource language pair (English-Telugu). 

\subsubsection{Fine-Tuning a Pre-Trained Model}
We utilize the \texttt{Helsinki-NLP/opus-mt-tc-big-en-it} \cite{tiedemann-thottingal-2020-opus} model, which is pre-trained on the English-Italian language pair. This model is fine-tuned on our English-Telugu dataset. Fine-tuning is carried out using key functions such as \texttt{Hugging Face’s Trainer API} and \texttt{train} from the transformer library.

To enhance the efficiency and performance of our fine-tuning process, we implement \textit{FlashAttention}, a novel attention mechanism that optimizes the standard attention mechanism used in transformer models. FlashAttention is an IO-aware exact attention algorithm that optimizes memory usage and improves computational efficiency in transformer models \cite{Dao2022FlashAttentionFA}. This approach reduces the number of memory reads and writes, making it particularly effective for long sequences.

The standard attention mechanism computes the attention scores using the following equation:

\[
\text{Attention}(Q, K, V) = \text{softmax}\left(\frac{QK^T}{\sqrt{d_k}}\right)V
\]

where:
- \( Q \) is the query matrix,
- \( K \) is the key matrix,
- \( V \) is the value matrix,
- \( d_k \) is the dimension of the key vectors.

FlashAttention introduces a more efficient way to compute this by using a technique known as \textit{softmax with efficient memory access}. The key equations for FlashAttention can be summarized as follows:

\textbf{1. Query-Key-Value Projections}: FlashAttention performs projections as follows:

   \[
   Q, K, V = \text{Linear}(X) \quad \text{for } X \in \mathbb{R}^{N \times d}
   \]

   where \( N \) is the sequence length and \( d \) is the dimensionality of the model.

\textbf{2. Scaled Dot-Product Attention}: The attention scores are calculated as:

   \[
   \text{Attention}(Q, K, V) = \text{softmax}\left(\frac{QK^T}{\sqrt{d}}\right)V
   \]

\textbf{3. Memory Efficiency}: FlashAttention optimizes the memory usage during the computation by ensuring that the gradients are computed in a manner that requires less memory and processing power.

\subsubsection{Advantages of FlashAttention in Our Project}

\begin{itemize}
    \item \textbf{Reduced Memory Usage:} FlashAttention's optimization techniques significantly decrease the memory requirements during training, allowing us to work with larger models or longer sequences without running into memory bottlenecks.
    
    \item \textbf{Faster Training Times:} By reducing the complexity of the attention calculations, FlashAttention speeds up the training process, enabling quicker iterations on model tuning and adjustments.
    
    \item \textbf{Improved Scalability:} The efficiency of FlashAttention makes it more feasible to scale our model training to larger datasets, which is particularly advantageous when dealing with synthetic data generated through backtranslation.
    
    \item \textbf{Maintained Performance:} Despite the improvements in efficiency, FlashAttention retains the performance of traditional attention mechanisms, ensuring that the translation quality remains high.
\end{itemize}

By incorporating FlashAttention into our fine-tuning process, we leverage these advantages to enhance the overall effectiveness and efficiency of our English-Telugu translation model.

\subsection{Backtranslation Process}

Backtranslation \cite{edunov2018} is a powerful technique employed to generate synthetic parallel data, which can be particularly beneficial in scenarios where there is a lack of sufficient parallel corpora. This method involves translating monolingual sentences from one language to another, and subsequently utilizing these translations to augment the existing dataset. The core idea behind backtranslation is to create additional training examples that can help improve the performance of a translation model, especially in low-resource language pairs.

\subsubsection{Generating Synthetic Data}
The backtranslation process begins with the translation of Telugu monolingual sentences into English. This is achieved using a reverse translation model specifically trained for this purpose. The synthetic English sentences generated through this translation process are then paired with their corresponding original Telugu sentences. By creating these additional pairs, we effectively increase the size and diversity of the dataset, thereby enhancing the robustness and generalization capabilities of the translation model.

To implement this process, a key function is utilized to perform the conversion of Telugu text into English. This function ensures that the translation maintains the semantic meaning of the original text while adapting it to the linguistic structure of the target language. Once the Telugu sentences are translated into English, the resulting tokenized sentences are incorporated into the training dataset. The generation of synthetic data is accomplished through the use of well-defined utility functions that handle the processes of tokenization, inference, and decoding. 

Tokenization is crucial because it transforms the text into a format suitable for model training, allowing the translation model to operate on subword units rather than entire sentences. Inference refers to the process of using the trained model to predict the translation of new input data, while decoding involves converting the model's output (token IDs) back into human-readable text.

\subsubsection{Bidirectional Backtranslation}
In addition to translating Telugu sentences into English, backtranslation can be applied in the reverse direction as well. This means that English monolingual sentences are also translated into Telugu. By employing a bidirectional approach, we can further enhance the quality of translations in both languages. This dual application of backtranslation ensures that the model is exposed to a more diverse set of training examples, improving its ability to handle various linguistic structures and idiomatic expressions.

The implementation of the bidirectional translation mechanism is facilitated through utility functions that are designed to manage both forward and reverse translations. This ensures that the model can learn from diverse sentence pairs, which is critical for improving translation quality. By training on a comprehensive dataset that includes translations in both directions, the model becomes more adept at capturing the nuances of each language, ultimately leading to better translation outcomes.

In summary, the backtranslation process is an effective strategy for augmenting the training dataset, leveraging synthetic data to improve the performance of translation models, particularly in low-resource settings. By incorporating both monolingual sentences from each language and utilizing a bidirectional approach, we can enhance the model's ability to generate accurate and fluent translations.

\subsection{Iterative Fine-Tuning}

Once the synthetic data has been generated via backtranslation, the next step involves the iterative fine-tuning of the translation model on the augmented dataset. This iterative fine-tuning is crucial for enhancing the model's performance, as it allows the model to learn from an increasingly diverse set of training examples. The process consists of multiple cycles of training, during which the model is exposed not only to the original parallel sentences but also to the newly generated backtranslated sentences.

\subsubsection{Iterative Training Process}
In each iteration of the fine-tuning process, the synthetic data generated from the backtranslation is incorporated into the existing training set. This results in a combined dataset that consists of both the original sentences and the synthetic translations. The model is then fine-tuned iteratively on this enriched dataset, which serves to reinforce its understanding of linguistic structures and contextual nuances present in both languages.

The iterative fine-tuning involves multiple rounds of training, each contributing to refining the model’s translation accuracy. The process is designed to be dynamic, allowing for adjustments based on the performance of the model after each iteration. As the training progresses, the model gains exposure to a wider range of sentence constructions and vocabulary, leading to improved translation capabilities.

For our specific implementation, this iterative fine-tuning process has been conducted for a total of \textbf{five} iterations. In each iteration, the model undergoes a training cycle where it learns from both the original data and the augmented synthetic data generated through backtranslation. This ensures that the model not only retains the linguistic patterns learned from the initial training but also adapts to the variations introduced by the synthetic sentences.

The implementation of the iterative fine-tuning process relies on a set of functions designed to manage various aspects of the training workflow. These functions handle crucial tasks such as loading the dataset, updating the model parameters, and managing the training loop. By automating these processes, we can maintain a systematic approach to training, allowing for efficient experimentation and optimization of the model.

Overall, the iterative fine-tuning process, particularly when repeated for five iterations, significantly enhances the model's ability to generalize and produce high-quality translations. As illustrated in Fig. 1, which provides an overview of the backtranslation and iterative fine-tuning process, each cycle of training reinforces the knowledge gained in previous iterations, ultimately leading to a more robust and accurate translation system.

\begin{figure}[ht]
    \centering
    \begin{tikzpicture}[
        node distance=1.0cm,
        every node/.style={draw, rectangle, rounded corners, align=center, minimum width=2cm},
        arrow/.style={-{Stealth[length=3mm]}, thick}
    ]

    \node (monolingual) {Collect Monolingual Sentences};
    \node (translate_telugu) [below=of monolingual] {Translate Telugu to English \\ (Using Reverse Model)};
    \node (pair_sentences) [below=of translate_telugu] {Pair Original Telugu \\ Sentences with Translations};
    \node (generate_synthetic) [below=of pair_sentences] {Generate Synthetic Data};
    \node (combine_data) [below=of generate_synthetic] {Combine Original and Synthetic Data};
    \node (iterative_finetune) [below=of combine_data] {Iterative Fine-Tuning \\ (5 Iterations)};
    \node (improved_model) [below=of iterative_finetune] {Improved Translation Model};

    \draw[arrow] (monolingual) -- (translate_telugu);
    \draw[arrow] (translate_telugu) -- (pair_sentences);
    \draw[arrow] (pair_sentences) -- (generate_synthetic);
    \draw[arrow] (generate_synthetic) -- (combine_data);
    \draw[arrow] (combine_data) -- (iterative_finetune);
    \draw[arrow] (iterative_finetune) -- (improved_model);

    \end{tikzpicture}
    \caption{Overview of the Backtranslation and Iterative Fine-Tuning Process}
    \label{fig:backtranslation_finetuning}
\end{figure}
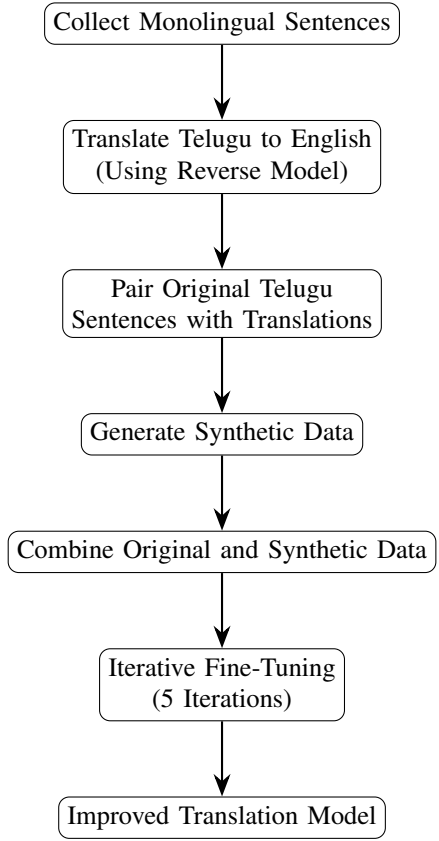
\subsection{Evaluation and Translation Quality}

\subsubsection{BLEU Score Evaluation}
To measure the effectiveness of the model, we employ the BLEU (Bilingual Evaluation Understudy) score \cite{papineni2002}. The BLEU score is calculated by comparing the model's translated sentences to one or more reference translations. It assesses the overlap of n-grams (contiguous sequences of n items from the translated text) between the generated translations and the reference translations. 

The calculation involves several steps:

1. \textbf{N-gram Precision:} For each n-gram (e.g., unigrams, bigrams, trigrams) in the candidate translation, we compute the precision, which is the ratio of the number of matching n-grams in the candidate translation to the total number of n-grams in the candidate translation. 

2. \textbf{Brevity Penalty:} To discourage overly short translations, a brevity penalty is applied if the candidate translation is shorter than the reference translations. The brevity penalty is calculated as:
   \[
   BP = 
   \begin{cases} 
   1 & \text{if } c > r \\
   e^{(1 - \frac{r}{c})} & \text{if } c \leq r 
   \end{cases}
   \]
   where \( c \) is the length of the candidate translation and \( r \) is the length of the closest reference translation.

3. \textbf{Final BLEU Score Calculation:} The final BLEU score is computed as a geometric mean of the n-gram precisions multiplied by the brevity penalty:
   \[
   \text{BLEU} = BP \cdot \exp\left(\sum_{n=1}^{N} w_n \log p_n\right)
   \]
   where \( p_n \) is the precision of n-grams and \( w_n \) is the weight assigned to each precision score (typically \( w_n = \frac{1}{N} \) for uniform weights).

In our implementation, the BLEU score is computed using evaluation libraries that support corpus-based scoring, allowing us to efficiently evaluate the translation quality of our model against multiple reference translations. A higher BLEU score indicates better translation quality, thus validating the effectiveness of our approach in enhancing the model's performance.

\section{Experimental Setup}

    \subsection{Training Configuration}
        The training is performed using Hugging Face’s Trainer API, with parameters set for batch size, learning rate, and number of epochs. Flash Attention is employed for memory efficiency.
    \subsection{Evaluation Metrics}
        The model is evaluated using the BLEU score, a widely accepted metric for machine translation tasks. A larger test dataset of 400 English sentences and their corresponding ground truth Telugu translations is used for evaluation.
  \begin{table}[ht]
        \centering
        \caption{Evaluation Metrics for English-to-Telugu Translation Model}
        \label{tab:evaluation_metrics}
        \begin{tabular}{|c|c|}
            \hline
            \textbf{Evaluation Metric} & \textbf{Value} \\ \hline
            BLEU Score En-Te (Train) & 14.85 \\ \hline
            BLEU Score En-Te (Test) & 11.69 \\ \hline
        \end{tabular}
    \end{table}

\section*{Results and Discussion}

The fine-tuned model achieved a BLEU score of 11.69 on the test dataset. The entire training and fine-tuning process took approximately 37 hours on an 8GB Nvidia GPU. 

Table I presents the evaluation metrics for the English-to-Telugu translation model, providing insights into the model's performance across various criteria. Additionally, Fig. 2 illustrates sample translations from English to Telugu, showcasing the model's capabilities in generating translations that maintain the context and meaning of the original sentences.


\begin{figure}
    \centering
    \includegraphics[width=0.85\linewidth]{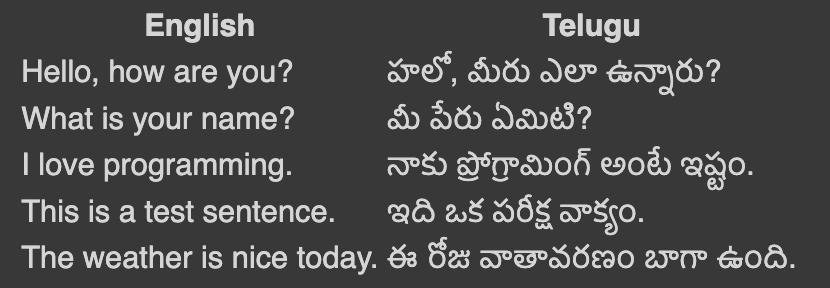}
    \caption{Sample Translations from English to Telugu}
    \label{fig:enter-label}
\end{figure}

\section{CONCLUSION AND FUTURE WORK}
This project illustrates the significant potential for developing an effective English-to-Telugu translator, even when faced with the challenges of limited data availability. By leveraging advanced transfer learning techniques, we have established a framework that allows us to adapt a pre-trained model to a low-resource language pair, enabling the utilization of insights gained from high-resource language pairs. This methodology not only enhances the initial capabilities of our translation model but also lays the groundwork for continual improvements.

While the current iteration of the model has yielded a low BLEU score, this score should not be interpreted as a failure; rather, it highlights the inherent difficulties in translating between languages with distinct grammatical structures, vocabularies, and cultural nuances. The low score emphasizes the need for ongoing research and refinement, suggesting that our efforts in transfer learning and iterative backtranslation serve as a robust starting point for future enhancements.

The integration of iterative backtranslation is particularly noteworthy, as it allows us to generate synthetic parallel data that enriches our training dataset. By translating monolingual sentences back and forth between English and Telugu, we can create diverse sentence structures and increase the volume of available data. This process is essential for improving the model’s understanding of the target language, thereby enhancing its translation capabilities.

Looking ahead, several avenues for improvement can be pursued. Firstly, enhancing the model architecture by exploring more advanced neural network designs, such as incorporating transformer-based architectures with FlashAttention, can lead to better performance and efficiency. This could involve experimenting with different configurations, such as varying the number of layers, attention heads, and embedding dimensions, to find the optimal setup for the English-Telugu translation task.

Secondly, optimizing training parameters, including learning rates, batch sizes, and dropout rates, is crucial. Fine-tuning these parameters through systematic experimentation can lead to improved convergence rates and ultimately better translation accuracy. Utilizing techniques such as learning rate scheduling or adaptive optimization algorithms could further enhance the training process.

Moreover, expanding the dataset to encompass a broader range of sentence structures, styles, and contexts will provide the model with a richer understanding of both languages. Incorporating domain-specific texts, idiomatic expressions, and culturally relevant phrases will be essential for creating a more nuanced and accurate translation system. Engaging with native speakers and linguistic experts can also offer invaluable insights into the subtleties of the Telugu language, which could be used to refine our training data.

In conclusion, while the current model demonstrates a promising foundation for English-to-Telugu translation, significant room for improvement remains. By pursuing the outlined strategies—enhancing model architecture, optimizing training parameters, and expanding the dataset—we can substantially elevate the performance of the translator. Such advancements will not only improve translation quality but will also increase the model’s applicability in practical settings, making it a valuable tool for communication and understanding between English and Telugu speakers. Through persistent effort and innovation, we can transform this project into a robust solution that effectively bridges linguistic divides.

\section*{Acknowledgment}

The author gratefully acknowledges the support provided by ABV-IIITM Gwalior.

\bibliographystyle{IEEEtran}
\bibliography{reference}

\begin{thebibliography}{10}
\providecommand{\url}[1]{#1}
\csname url@samestyle\endcsname
\providecommand{\newblock}{\relax}
\providecommand{\bibinfo}[2]{#2}
\providecommand{\BIBentrySTDinterwordspacing}{\spaceskip=0pt\relax}
\providecommand{\BIBentryALTinterwordstretchfactor}{4}
\providecommand{\BIBentryALTinterwordspacing}{\spaceskip=\fontdimen2\font plus
\BIBentryALTinterwordstretchfactor\fontdimen3\font minus \fontdimen4\font\relax}
\providecommand{\BIBforeignlanguage}[2]{{%
\expandafter\ifx\csname l@#1\endcsname\relax
\typeout{** WARNING: IEEEtran.bst: No hyphenation pattern has been}%
\typeout{** loaded for the language `#1'. Using the pattern for}%
\typeout{** the default language instead.}%
\else
\language=\csname l@#1\endcsname
\fi
#2}}
\providecommand{\BIBdecl}{\relax}
\BIBdecl

\bibitem{gala2023indictrans}
\BIBentryALTinterwordspacing
J.~Gala, P.~A. Chitale, R.~AK, V.~Gumma, S.~Doddapaneni, A.~Kumar, J.~Nawale, A.~Sujatha, R.~Puduppully, V.~Raghavan, P.~Kumar, M.~M. Khapra, R.~Dabre, and A.~Kunchukuttan, ``Indictrans2: Towards high-quality and accessible machine translation models for all 22 scheduled indian languages,'' 2023. [Online]. Available: \url{https://doi.org/10.48550/arXiv.2305.16307}
\BIBentrySTDinterwordspacing

\bibitem{bahdanau2014}
D.~Bahdanau, K.~Cho, and Y.~Bengio, ``Neural machine translation by jointly learning to align and translate,'' in \emph{Proceedings of the International Conference on Learning Representations (ICLR)}, 2015.

\bibitem{kudo2018sentencepiece}
\BIBentryALTinterwordspacing
T.~Kudo and J.~Richardson, ``Sentencepiece: A simple and language independent subword tokenizer and detokenizer for neural text processing,'' 2018. [Online]. Available: \url{https://doi.org/10.48550/arXiv.1808.06226}
\BIBentrySTDinterwordspacing

\bibitem{sennrich2016}
R.~S. et~al., ``Improving neural machine translation models with monolingual data,'' in \emph{Proceedings of ACL}, 2016, pp. 83 --91.

\bibitem{edunov2018}
S.~Edunov, M.~Ott, M.~Auli, and D.~Grangier, ``Understanding back-translation at scale,'' \emph{Proceedings of the Conference on Empirical Methods in Natural Language Processing (EMNLP)}, pp. 489--495, 2018.

\bibitem{vaswani2017}
A.~Vaswani, N.~Shard, N.~Parmar, J.~Uszkoreit, L.~Jones, A.~N. Gomez, Å.~Kaiser, K.~Kattner, and N.~P. et~al., ``Attention is all you need,'' in \emph{Advances in Neural Information Processing Systems (NeurIPS)}, 2017.

\bibitem{bawden-etal-2020-university}
\BIBentryALTinterwordspacing
R.~Bawden, A.~Birch, R.~Dobreva, A.~Oncevay, A.~V.~M. Barone, and P.~Williams, ``The university of edinburgh's english-tamil and english-inuktitut submissions to the wmt20 news translation task,'' in \emph{Proceedings of the Fifth Conference on Machine Translation}.\hskip 1em plus 0.5em minus 0.4em\relax Scotland: School of Informatics, University of Edinburgh, 2020. [Online]. Available: \url{https://www.statmt.org/wmt20/pdf/2020.wmt-1.38.pdf}
\BIBentrySTDinterwordspacing

\bibitem{sennrich2015}
R.~Sennrich, B.~Haddow, and A.~Birch, ``Neural machine translation of rare words with subword units,'' in \emph{Proceedings of the Association for Computational Linguistics (ACL)}, 2016, pp. 1715--1725.

\bibitem{tiedemann-thottingal-2020-opus}
\BIBentryALTinterwordspacing
J.~Tiedemann and S.~Thottingal, ``{OPUS}-{MT} {--} building open translation services for the world,'' in \emph{Proceedings of the 22nd Annual Conference of the European Association for Machine Translation}.\hskip 1em plus 0.5em minus 0.4em\relax Lisboa, Portugal: European Association for Machine Translation, Nov. 2020, pp. 479--480. [Online]. Available: \url{https://aclanthology.org/2020.eamt-1.61}
\BIBentrySTDinterwordspacing

\bibitem{Dao2022FlashAttentionFA}
\BIBentryALTinterwordspacing
T.~Dao, D.~Y. Fu, S.~Ermon, A.~Rudra, and C.~R'e, ``Flashattention: Fast and memory-efficient exact attention with io-awareness,'' \emph{ArXiv}, vol. abs/2205.14135, 2022. [Online]. Available: \url{https://arxiv.org/abs/2205.14135}
\BIBentrySTDinterwordspacing

\bibitem{papineni2002}
K.~Papineni, S.~Roukos, T.~Ward, and W.-J. Zhu, ``Bleu: a method for automatic evaluation of machine translation,'' \emph{Proceedings of the 40th Annual Meeting of the Association for Computational Linguistics (ACL)}, pp. 311--318, 2002.

\end{thebibliography}

\end{document}